%% file: main.tex
\documentclass[lettersize,journal]{IEEEtran} 

\IEEEoverridecommandlockouts                              

\usepackage{graphicx}
\usepackage{amsmath,amssymb}
\usepackage{cite}
\usepackage{float}
\usepackage[caption=false,font=footnotesize]{subfig}
\usepackage{algorithm}
\usepackage{algorithmic}
\usepackage{color}
\definecolor{review_color}{rgb}{0,0,0}
\definecolor{orange}{rgb}{0,0,0}
\definecolor{modif}{rgb}{0,0,0}
\definecolor{gray}{rgb}{0.5,0.5,0.5}
\graphicspath{{img/},{fig/}}

\begin{document}


\title{
Soft Two-degree-of-freedom Dielectric Elastomer Position Sensor Exhibiting Linear Behavior
}

\author{Alexandre Girard$^1$, Jean-Philippe Lucking Bigu\'{e}$^1$, Benjamin M. O'Brien$^2$, Todd A. Gisby$^2$, Iain A. Anderson$^2$ and Jean-S\'{e}bastien Plante$^1$%
\thanks{This work was supported by the Natural Sciences and Engineering Research Council of Canada (NSERC) and the Fonds Qu\'{e}b\'{e}cois de la Recherche sur la Nature et les Technologies (FQRNT).}
\thanks{$^1$A. Girard, J.-P. Lucking Bigu\'{e} and J.-S. Plante are with the Department of Mechanical Engineering, Universit\'{e} de Sherbrooke, Sherbrooke, Qu\'{e}bec, Canada}%
\thanks{$^2$Ben O'Brien, Todd Gisby and Iain Anderson are with the Biomimetics Lab, Auckland Bioengineering Institute, The University of Auckland, Auckland}
\thanks{$^3$ \textcopyright IEEE. Personal use of this material is permitted. Permission from IEEE must be obtained for all other uses, in any current or future media, including reprinting/republishing this material for advertising or promotional purposes, creating new collective works, for resale or redistribution to servers or lists, or reuse of any copyrighted component of this work in other works. DOI:10.1109/TMECH.2014.2307006}
}

\markboth{IEEE/ASME Transactions on Mechatronics,~Vol.~20, No.~1,February~2015, Preprint version. DOI:10.1109/TMECH.2014.2307006$^3$}{}

\maketitle

\begin{abstract}
Soft robots could bring robotic systems to new horizons, by enabling safe human-machine interaction. For precise \textcolor{orange}{control, these }soft structures require high level position feedback that is not easily achieved through conventional \textcolor{modif}{one-degree-of-freedom} (DOF) sensing apparatus. In this paper, a soft \textcolor{modif}{two-DOF} dielectric elastomer (DE) sensor is specifically designed to provide accurate position feedback for a soft polymer robotic manipulator. The technology is exemplified on a soft robot intended for MRI-guided prostate interventions. DEs are chosen for their major advantages of softness, high strains, low cost and embedded \textcolor{modif}{multiple-DOF} \textcolor{modif}{sensing} capability, providing excellent system integration. A geometrical model of the proposed DE sensor is developed and compared to experimental results in order to understand sensor mechanics. Using a differential measurement approach, a handmade prototype provided linear sensory behavior and 0.2 mm accuracy on \textcolor{modif}{two-DOF}. This correlates to a 0.7\% error over the sensor's 30 mm x 30 mm planar range, demonstrating the outstanding potential of DE technology for accurate multi-DOF position sensing.
\end{abstract}


\input{sections/intro}

\input{sections/design}

\input{sections/experiments}

\input{sections/discussion}

\input{sections/conclusion}

\input{sections/ending}

\bibliographystyle{IEEEtran}
\bibliography{main}

\end{document}

%% file: sections/intro.tex
\section{Introduction}


\IEEEPARstart{M}{odern} robotic systems are generally confined to restricted areas, since they are unable to safely interact with uncertain environments. \textcolor{modif}{Robots} using geared electric motors cannot safely grasp an object, walk on uneven ground and interact with a human or another robot due to their high inertia and stiff position-controlled actuators \cite{fauteux_dual-differential_2010}. In contrast, soft robots possess the inherent advantages of lightness and compliance, which \textcolor{modif}{allow them} to perform a wide variety of tasks requiring safe interaction. A promising example of soft robot architecture is an all-polymer body, actuated by pneumatic cells \cite{shepherd_multigait_2011}\cite{festo_bionic_2010}. Such robots are cheap to manufacture (molded structure), very robust (low number of moving parts) and shock resistant (compliant structure), in addition to enable safe \textcolor{orange}{human-machine} interactions. 

\begin{figure}[t]
	\centering
		\subfloat[Rigid]{
		\label{fig:rigidrobot}
		\includegraphics[width=0.22\textwidth]{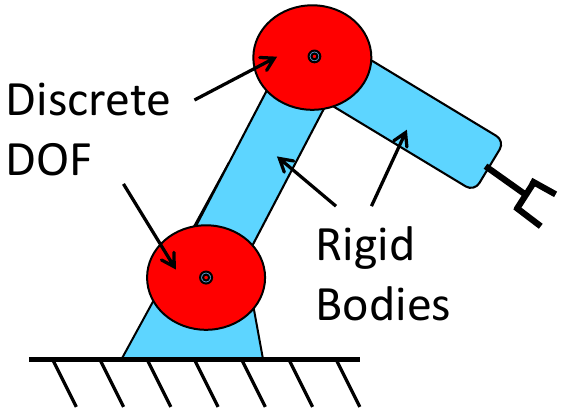}}
  	~
  	\subfloat[Soft]{
  	\label{fig:softrobot}
  	\includegraphics[width=0.22\textwidth]{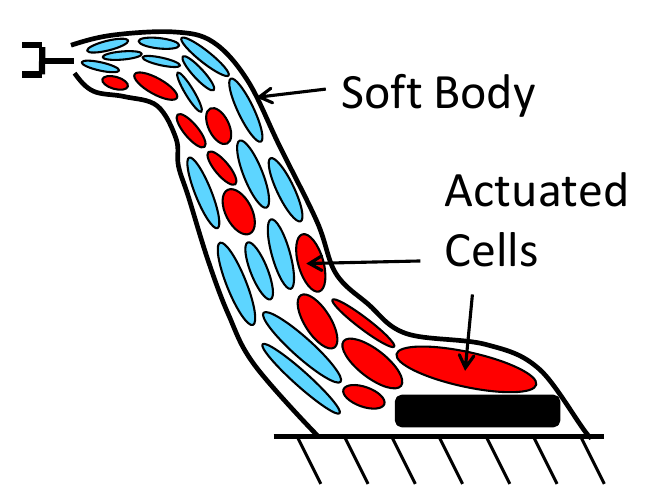}}
 	  
  	\caption{\textcolor{orange}{Rigid robotics (a) vs. soft robotics (b)}}
	\label{fig:robotics}
	\vspace{-10pt}
\end{figure}

As shown in Fig. \ref{fig:robotics}, classical (rigid) robots are composed of discrete degrees-of-freedom (DOF) linked together by rigid bodies (Fig. \ref{fig:rigidrobot}), while soft robots are composed of a plurality of actuation cells, embedded in a soft body (Fig. \ref{fig:softrobot}). The rigid body architecture relies on a fundamental package of joint-actuator-sensor for each discrete DOF. \textcolor{orange}{To reconstruct the end-effector position, local one-DOF sensor measurements are combined using kinematical models of the rigid structure.} In soft body architectures, the end-effector cannot be constrained by rigid bodies, thus requiring compliant sensors. Moreover, \textcolor{modif}{as} soft robots typically possess more internal DOF than the end-effector's six spatial constraints\textcolor{modif}{, reconstructing} the end-effector position using multiple local soft sensors requires complex models that link local strains to global deformations. A more practical approach, used in this paper, measures the end-effector's position directly through multi-DOF soft sensors, resulting in compact and integrated \textcolor{orange}{robotic systems.}

\subsection{Background}



\begin{figure}[h!]
	\centering
		\subfloat[Initial state]{
		\label{fig:DEA_c1}
		\includegraphics[width=0.22\textwidth]{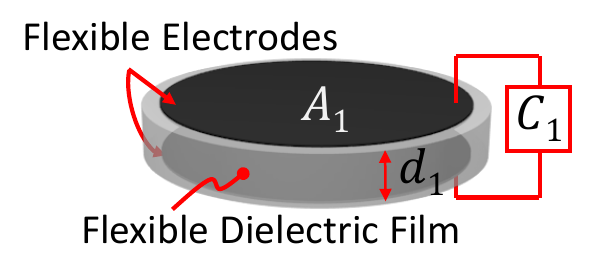}}
  	~
  	\subfloat[Deformed state]{
  	\label{fig:DEA_c2}
  	\includegraphics[width=0.22\textwidth]{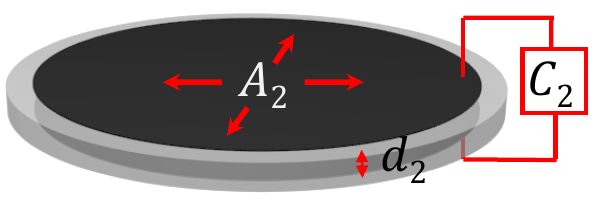}}
 	  
  	\caption{DEs are flexible capacitors}
	\label{fig:DEA_operation}
\end{figure}

As high strains and low stiffness are required for soft sensing, most promising technologies are based on flexible polymers. Dielectric elastomers (DE) are particularly interesting for their high strain capability as well as their embedded design potential and soft matrix. As illustrated in Fig. \ref{fig:DEA_operation}, DE technology is best represented by a flexible capacitor, composed of a pair of electrodes, painted on either side of a soft dielectric membrane. When a DE film is deformed by an external load, its area $A$ and thickness $d$ are modified, causing a variation of the underlying capacitance $C$ in accordance with the flat capacitor equation:
%
\begin{align}
C=\epsilon_{0}\epsilon_{r}\frac{A}{d}
\label{eq:capacitance}
\end{align}
where $\epsilon_{0}$ is the permittivity of freespace and $\epsilon_{r}$ is the relative dielectric constant of the film. In comparison with similar operating principles, such as piezoresistivity that offer unpredictable behavior and large hysteresis, the high stain piezocapacitive operating principle has shown highly accurate and repeatable \textcolor{orange}{behaviour}, as the capacitance is a purely geometric property \cite{ulmen_robust_2010}\cite{cohen_highly_2012}.

\textcolor{orange}{The operating principle shown in Fig. \ref{fig:DEA_operation} has been used for many soft robotic purposes, ranging from actuation and energy harvesting, to soft logic and sensing \cite{anderson_multi-functional_2012}. \textcolor{modif}{In terms of accuracy, a} self-sensing DE actuator showed a 10 $\mu$m static and 50 $\mu$m dynamic accuracy over a 0.5 mm stroke \cite{jung_self-sensing_2008}. \textcolor{modif}{A} fiber reinforced self-sensing McKibben muscle was also created by placing a dielectric elastomer sensor in direct contact with the muscle's inner surface \cite{son_finite_2009}. Experimental results showed a linear capacitance-stretch relationship when no pressure was applied. Pressure-induced non-linearities were modeled, providing an accuracy of 6\% when compared to experimental results for strains up to 50\%.} Self-sensing was also used in the closed loop control of a planar DE actuator \cite{gisby_closed_2011}, which could simultaneously actuate a tactile finger and sense the characteristics of contact between the finger and an object \cite{gisby_how_2012}. Although promising, these actuator-sensor devices focus on local one DOF measurements such as used in rigit robot architectures, and much further development is required to extend this work to global positioning of soft robots.

\textcolor{orange}{For pure sensing applications, DE technology has raised interest for monitoring human movement \cite{pelrine_electroactive_2004}, infrastructures health \cite{laflamme_soft_2012} and blood pressure\cite{iskandarani_sensing_2012}.} Flexible artificial skin has also been developed using both piezoresistive\cite{park_design_2012} and piezocapacitive\cite{ulmen_robust_2010} effects, demonstrating multi-DOF tactile feedback in single embedded soft devices. 
Proof-of-principle \textcolor{modif}{of} DE motion control was demonstrated through the development of a one\textcolor{modif}{-}DOF strain gauge \cite{cohen_highly_2012}. The capacitance-based prototype showed linear, stable and reliable strain-capacitance behavior, as compared to similar devices that rely on resistivity variations. Expanding such characteristics to a global multi-DOF soft sensor, such as proposed in this paper, could provide better accuracy \textcolor{orange}{than local approaches}, by avoiding complex coupling with actuators inputs and providing linear capacitance-displacement relations that can be easily calibrated and measured.

 
\subsection{Approach and Results}

This paper demonstrates the feasibility and potential of using integrated, multi-DOF, DE sensors for global position feedback of soft robots by developing an application example on a soft robotic system destined for MRI-guided interventions.  In section \ref{sec:design}, the robotic system is described with its sensing requirements, \textcolor{modif}{while} the proposed sensing system solution, consisting of a pair of identical 2 DOF sensors, is presented and studied using analytical models. The static and dynamic sensing results of a 2 DOF prototype then follow in section \ref{sec:exp}. Overall, the prototype \textcolor{modif}{shows} excellent results when used in combination with a differential measurement approach, demonstrating 0.2 mm accuracy on both DOF.




%% file: sections/design.tex
\section{Design \textcolor{orange}{and modeling}}
\label{sec:design}

\subsection{Robotic System}

\begin{figure}[ht]
	\centering
		\subfloat[Gobal view]{
		\label{fig:robotA}
		\includegraphics[width=0.26\textwidth]{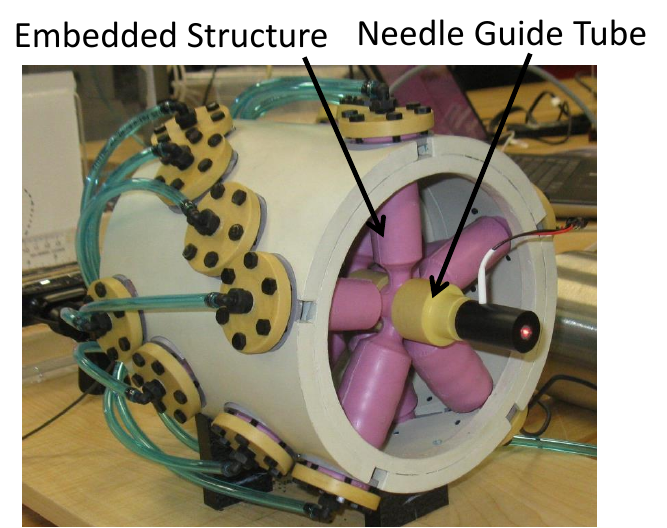}}
  	~
  	\subfloat[Internal soft structure]{
  	\label{fig:robotB}
  	\includegraphics[width=0.177\textwidth]{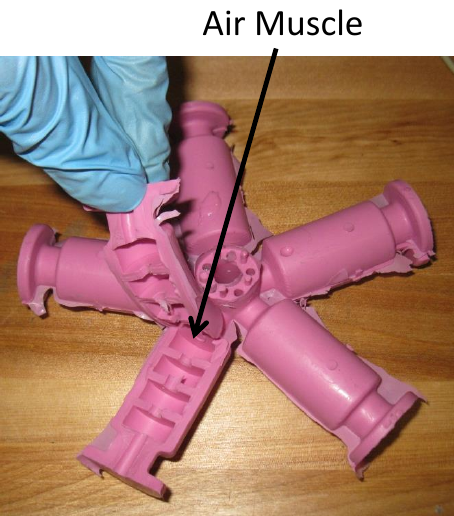}}
 	  
  	\caption{Soft robot for MRI-guided interventions \cite{miron_design_2013}}
	\label{fig:robot}
\end{figure}

The soft robotic system, for which the sensor is developed, is presented on Fig. \ref{fig:robot}. The purpose of this robot is to orient a needle guide inside a magnetic resonance imaging (MRI) \textcolor{orange}{scanner }for accurate medical interventions \cite{miron_design_2013}.

\begin{figure}[ht]
	\centering
		\includegraphics[width=0.48\textwidth]{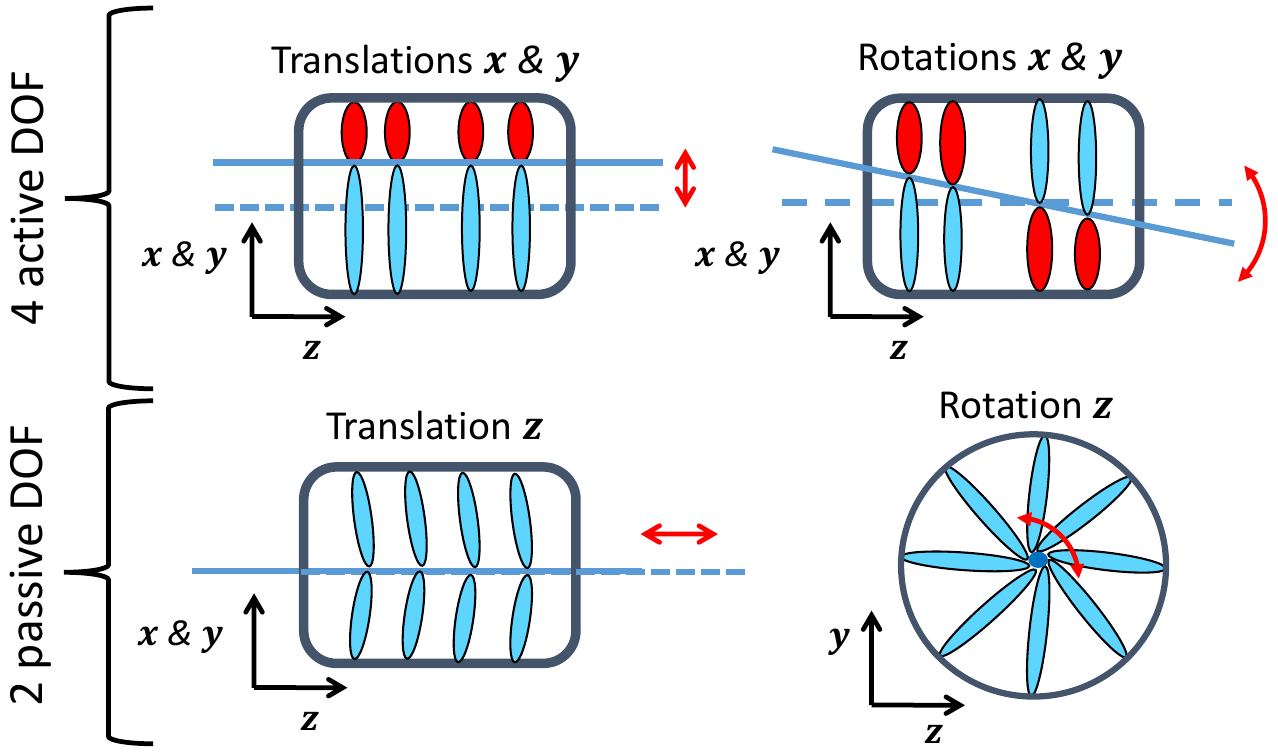}
	\caption{\textcolor{orange}{Robot six degrees-of-freedom. On top, translations and rotations of the needle guide, about the $x$ and $y$ axis, controlled by muscles inflations (active DOF). Below, translation and rotation, about the $z$ axis, which may be caused by external forces (passive DOF).}}
	\label{fig:DOF}
\end{figure}

The soft robot has 20 internal DOF, provided by 20 binary air muscles placed in a parallel configuration around a rigid needle guide. As shown in Fig. \ref{fig:DOF}, this configuration yields four active (controllable) DOF and two passive, yet soft, DOF.

\subsection{Sensor Requirement}
\label{sec:requirement}

The main requirement for the DE sensor is to accurately sense the four active DOF, \textcolor{orange}{shown in Fig. \ref{fig:DOF}}, without constraining any of the six global DOF of the system. The sensor must also be compatible with the geometric constraints of the robot, the motion amplitudes and the high intensity magnetic field of MRI. The resolution of the MRI images sets the precision requirement at $\pm$ 1mm at the needle-tip, located 200 mm away from the robot. To achieve this accuracy and avoid complex modeling, it is also desirable to develop a sensor that is decoupled from the actuation system (i.e. muscle inflation). Finally, a simple design that uses a low number of moving parts will be preferred, in order to keep the advantages (i.e. robustness and low cost) of embedded soft robotic systems. Note that the bandwidth requirement is almost inexistent for this specific application, as the robot is designed to be a point-to-point static positioner. 

\subsection{Proposed Design}

\begin{figure}[htbp]
	\centering
		\subfloat[Lateral view of the system]{
		\label{fig:solutionA}
		\includegraphics[width=0.20\textwidth]{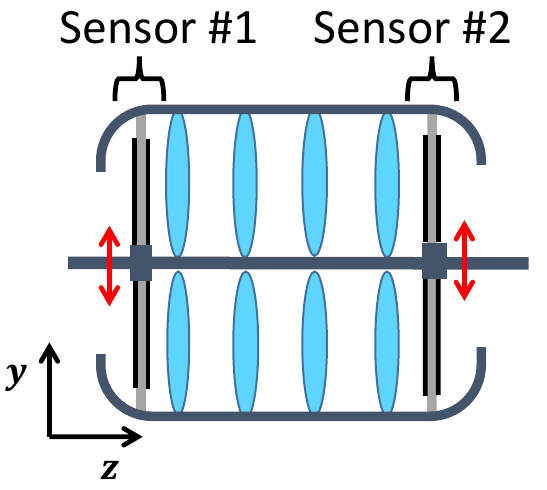}}
  	~
  	\subfloat[Front view of one sensor unit]{
  	\label{fig:solutionB}
  	\includegraphics[width=0.25\textwidth]{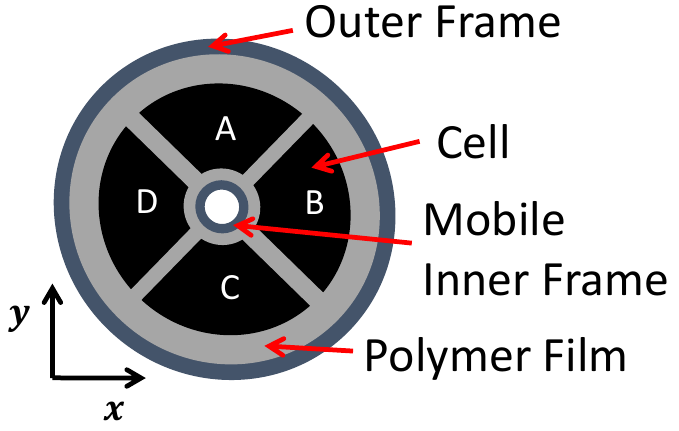}}
 	  
  	\caption{Four DOF sensing solution with two planar sensors}
	\label{fig:solution}
\end{figure}

The proposed sensing approach combines a pair of identical two DOF planar sensors, fixed on both extremities of the robot, as seen in Fig. \ref{fig:solutionA}. Each two-DOF sensor are used to measure the in-plane \textcolor{orange}{($x$,$y$)} translations of the needle-guide, which is fixed to the mobile inner frames of the sensors. \textcolor{orange}{Each sensor use four cells (A, B, C and D), i.e. flexible capacitances formed by pairs of electrodes, located on either side of the dielectric film (see Fig. \ref{fig:solutionB}). Cells A and C measure displacements in the $y$ direction, while cells B and D measure displacements in the $x$ direction. The 3D position of the needle guide will then be reconstruted using the four measured diplacements ($x$ and $y$ displacements in both planes).}


While position measurements could be provided by sensing each of the four cells independently, the two opposing cell-per-DOF configuration, shown in Fig. \ref{fig:solutionB}, also enables differential measurement. This strategy, further detailed in section \ref{sec:diff}, will be shown to provide a linear sensor behavior (capacitance variations vs. displacement) and to cancel out the effects of parasitic motions (i.e. out-of-plane translation, out-of-plane rotation and in-plane rotation). Moreover differential measurements also provides robustness towards environment effects, such as temperature variations \cite{laflamme_soft_2012}.


\input{sections/analysis}

%% file: sections/analysis.tex
\subsection{Analysis}

In this section, both the \textcolor{orange}{one-cell and a two-cell model (differential approach)} are derived. The geometrical design of the sensor is then optimi\textcolor{modif}{z}ed in regards to the constraints imposed in section \ref{sec:requirement} by analysing the overall behavior and sensitivity of the capacitance-displacement relationship, as well as the influence of parasitic motion \textcolor{orange}{(all motion that are not pure in-plane ($x$,$y$) translations for both sensor inner frames, such as induced by the passive DOF).}

\subsubsection{One-cell Model}
\label{sec:Modelisation}

\begin{figure}[htbp]
	\centering
		\includegraphics[width=0.22\textwidth]{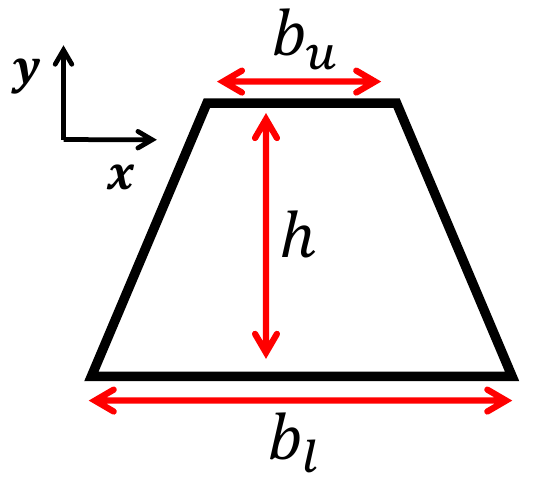}
	\caption{Geometric variables of the trapezoid electrode shape}
	\label{fig:trapz}
\end{figure}

To obtain simple design parameters, \textcolor{orange}{the geometry of an arbitrary electrode cell (here shown as electrode C) is approximated} by a trapezoid shape such as shown in Fig. \ref{fig:trapz}.
The capacitance $C$ of a cell can be related to its geometric and dielectric parameters, using eq. \eqref{eq:capacitance}:
\begin{align}
	C = \epsilon_r \epsilon_0 \frac{bh}{d}
\end{align}
where $b$ is the mean base $(b_u+b_l)/2$ of the trapezoid and $h$ is its height. The upper base $b_u$ and lower base $b_l$ are fixed on the rigid bodies of the inner and outer frames, thus remaining constants. The area of the cell will then only change for displacements of the inner frame that affect its height $h$, in other words, with $y$ axis motions.

The high bulk modulus of most polymers make them incompressible. Thus, when a cell is streched, its thickness will change, while the volume $V$ remains constant:
\begin{equation}
\label{eq:vcste}
V=bhd=cst.
\end{equation}
With the constant volume, the capacitance of a cell can be solely expressed as a fonction of its height :
\begin{equation}
C = k  h^2 
\label{eq:ch2}
\end{equation}
\begin{equation}
 k = \epsilon_r \epsilon_0 \frac{ b}{h_0 d_0} = cst. 
\end{equation} 
where $h_0$ and $d_0$ are the initial height and tickness of a cell.

\begin{figure}[ht]
	\centering
		\subfloat[$y$ axis displacement]{
		\label{fig:strain_a}
		\includegraphics[width=0.22\textwidth]{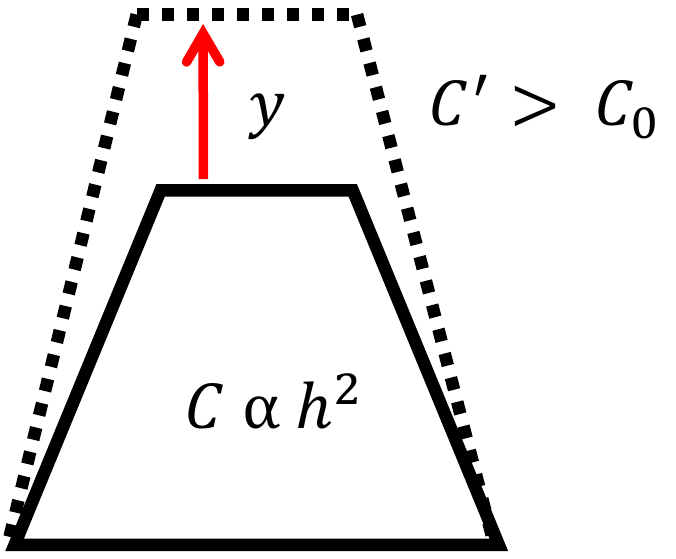}}
  	~
  	\subfloat[$x$ axis displacement]{
  	\label{fig:strain_b}
  	\includegraphics[width=0.18\textwidth]{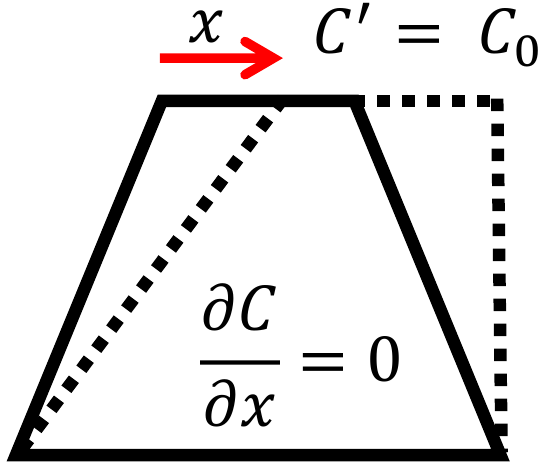}}
 	  
  	\caption{Capacitance changing with inner frame translations}
	\label{fig:strain}
\end{figure}

According to eq. \eqref{eq:ch2}, the capacitance $C$ of a cell varies with the square of its height $h$, for a displacement of the inner frame in the $y$ axis, as shown in Fig. \ref{fig:strain_a}. Note however, that the capacitance does not change for shear-like deformations, such as shown in Fig. \ref{fig:strain_b} where the inner frame moves in the $x$ direction.

The sensitivity of one cell can be derived from eq. \ref{eq:ch2} :  
\begin{align}
\frac{\partial C}{\partial y} &= \frac{2 \epsilon_r \epsilon_0 b}{d_0} \left( 1 + \frac{y}{h_0} \right) \label{eq:sens} \\
\frac{\partial C}{\partial x} &= 0 
\end{align}
where $y$ is the vertical displacement of the inner frame, equal to $y=h-h_0$, and $x$ the horizontal shear-like displacement of the inner frame.
As it is desirable to maximi\textcolor{modif}{z}e sensitivity, eq. \eqref{eq:sens} suggests to maximi\textcolor{modif}{z}e the dielectric constant $\epsilon_r$ of the material and the mean base $b$, while minimizing the initial thickness $d_0$. The initial height $h_0$ of a cell does not affect the sensitivity in first order, but a large $h_0$ value will minimize the non-linear behavior \textcolor{modif}{(see eq.\eqref{eq:sens})}.

\subsubsection{Two-cell Model}
\label{sec:diff}

\begin{figure}[htbp]
	\centering
		\includegraphics[width=0.28\textwidth]{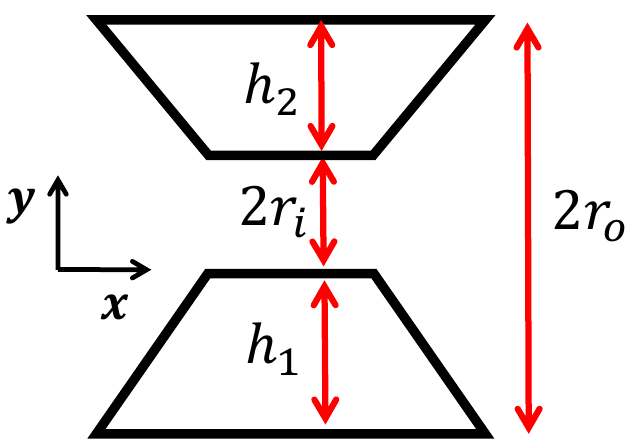}
	\caption{Two cells face to face for differential measurements}
	\label{fig:cellulefaceface}
\end{figure}

In this section, the capacitance difference between two opposing cells, \textcolor{orange}{as shown in Fig. \ref{fig:cellulefaceface},} is analysed. \textcolor{orange}{The variables $r_i$ and $r_o$ are respectively the inner and outer radius of the sensor.}

In this case, cell heights $h_i$ can be expresses as the combination of their initial height and the \textcolor{orange}{inner frame} $y$ displacement:
\begin{gather}
	h_1 = h_{0,1} + y \\
	h_2 = h_{0,2} - y
\end{gather}

The capacitance difference $\Delta C$ of two facing cells, is then expressed by :
\begin{align}
	\Delta C &= C_1 - C_2 \\
	&= k_1(h_{0,1} + y)^2 - k_2(h_{0,2} - y)^2 \\
	&= 2\left(k_1 h_{0,1} + k_2 h_{0,2}\right)y + \left(k_1 - k_2\right)y^2 + \Delta C_0
	\label{eq:DC}
\end{align}
where $\Delta C_0 = C_{0,1}-C_{0,2}$ is the initial capacitance difference. 

If the two cells are identical, i.e. $k_1 = k_2 = k$ and $h_{0,1} = h_{0,2} = h_0$, only the first term of eq. \eqref{eq:DC} remains, and the difference of capacitances becomes :
\begin{align}
	\Delta C &= 4 k h_0 y =  \frac{4 \epsilon_r \epsilon_0 b}{d_0} y
	\label{eq:dclin}
\end{align}
Thus the relation between the $y$ displacement and the difference of capacitance is linear for two identical cells facing each other. This is the key attribute of the differential sensing approach, as linearity greatly simplifies calibration, improves overall precision and would also be advantageous for computing-power limited system with numereous sensing cells.

In pratice, opposing cells will never be exactly identical due to manufacturing process uncertainties. This could cause an initial capacitance offset, $\Delta C_0$, that can easily be identified and compensated for. The non-linear second term of eq. \eqref{eq:DC}, $\left(k_1 - k_2\right)y^2$, is less desirable. To minimize its effect, the constant $k_i$ of both cells should be as close as possible. 

To further understand the implications of manufacturing errors on $b$ and $h_0$ (keeping identical film initial thickness $d_0$), the following relation for capacitance variation can be found:
\begin{align}
	\Delta C &= \frac{2 \epsilon_r \epsilon_0}{d_0} \left[ \left(b_1 + b_2\right)y + \left(\frac{b_1}{h_{0,1}} - \frac{b_2}{h_{0,2}}\right)y^2 \right] + \Delta C_0
	\label{eq:DC2}
\end{align}
The sensitivity of the differential approach is then :
\begin{align}
\frac{\partial \Delta C}{\partial y} &= \frac{4 \epsilon_r \epsilon_0}{d_0} \left[ \frac{b_1 + b_2}{2} + \left(\frac{b_1}{h_{0,1}} - \frac{b_2}{h_{0,2}}\right)y \right] 
\label{eq:sens2}
\end{align}
Eq. \eqref{eq:sens2} shows that to minimize the non-linearity in the presence of manufacturing errors, the cells should have high aspect ratio ($h_0/b$), thus minimizing the second non-linear term. Also, comparing eq. \eqref{eq:sens} with eq. \eqref{eq:sens2} shows that the differential approach is twice as sensitive as the one-cell measurement, while the design parameters have the same influence on the sensitivity of both cases.


\subsubsection{Parasitic Motion}
\label{sec:parasite}

The proposed DE sensors are \textcolor{orange}{at least an order of magnitude} less rigid than the polymer robot shown in Fig. \ref{fig:robot} and thus do not constrain the system on any DOF. As the robot can move in ways that will create out of plane motion, this section briefly investigates the impact of these parasitic motion\textcolor{modif}{s} on the capacitance readings and evaluates ways to minim\textcolor{modif}{ize} their \textcolor{modif}{impact} on in plane motion readings.

\begin{figure}[ht]
	\centering
		\includegraphics[width=0.45\textwidth]{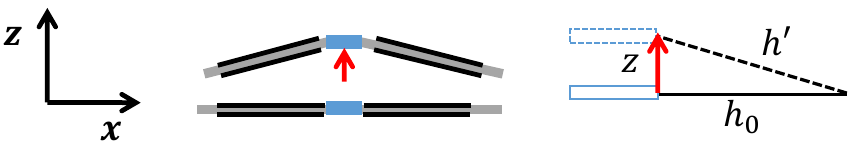}
	\caption{Out of plane translation along $z$ Axis}
	\label{fig:tz}
\end{figure}

\begin{figure}[ht]
	\centering
		\includegraphics[width=0.45\textwidth]{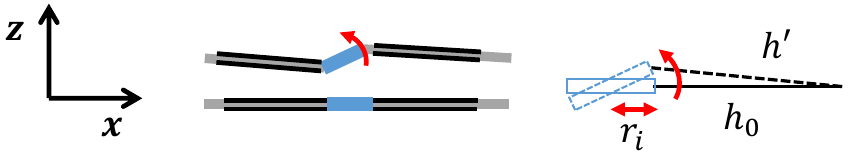}
	\caption{Out of plane rotation (same effect for rotation about $x$ and $y$ axis)}
	\label{fig:rxy}
\end{figure}

\begin{figure}[ht]
	\centering
		\includegraphics[width=0.45\textwidth]{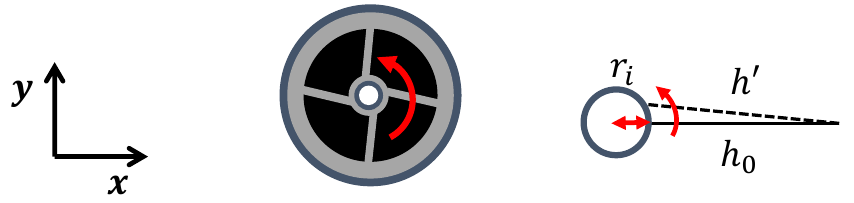}
	\caption{In plane rotation about $z$ axis}
	\label{fig:rz}
\end{figure}

Parasitic motion for one sensor includes translation in the $z$ direction (Fig. \ref{fig:tz}), as well as rotations about $x$ and $y$ axis (Fig. \ref{fig:rxy}) and about $z$ axis (Fig. \ref{fig:rz}). All the parasitic motions, when starting from the initial position of a sensor, are symmetric; thus their effects on two facing cells will be equivalent and canceled out by the differential mesurement. However for a combination of motions on many DOF, the parasitic effects are not perfectly symmetric. Hence, the sensor should still be designed to minimize the impact of parasitic motion because the differential measurement does not cancel their higher order effects. 

For a $z$ translation, all cells will be stretched. The height augmentation $\Delta h$ of the cells is given by :
\begin{equation}
\Delta h = h' - h  = \sqrt{h^2+z^2}-h
\label{eq:tz}
\end{equation}
Thus, in order to minimize this effect the initial height $h_0$ of cells should be as large as possible.

For an out of plane rotation, about the $x$ or $y$ axis, all cells are also going to be streched. The height augmentation $\Delta h$ of the cells is given by :
\begin{equation}
	\Delta h = \sqrt{(h+r_i)^2+r_i^2-2(h+r_i)r_i\cos(\theta)}-h
	\label{eq:rxy}
\end{equation}
Thus, in order to minimize the effet of rotations, the initial height $h_0$ of the cells should also be as large as possible and moreover the inner frame radius $r_i$ should be as small as possible. 

For small in plane rotation, cell capacitance should not change because the cell area remains constant. As with out of plane rotations, a tall initial height $h_0$ and a small inner frame radius $r_i$ will minimize the deformation caused by rotation around the $z$ axis. 

In all, the parasitic analysis shows that the initial height $h_0$ should be maximized while the inner radius $r_i$ must be minimized in order to minimize the parasitic motion degradation effets on in plane motion readings.

\subsubsection{Geometrical Design}

\begin{figure}[htb]
	\centering
		\includegraphics[width=0.32\textwidth]{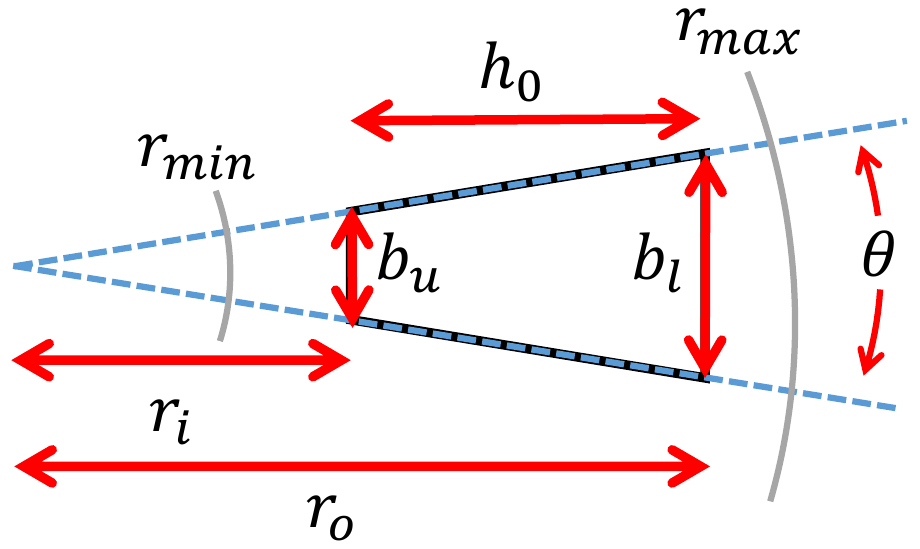}
	\caption{Optimization of the cell shape in a fraction of a circle}
	\label{fig:incircle}
\end{figure}

Using the guidelines of the last two sections, it is now possible to determine the best geometry for the sensor cells, starting with a trapezoid inscribed in a circular sector, shown in Fig. \ref{fig:incircle}.

To maximi\textcolor{modif}{z}e sensitivity to in-plane translation, the two bases ($b_u$ and $b_l$) should be as large as possible (see eq. \eqref{eq:dclin}). Hence the section angle $\theta$ is set to 90$^{\circ}$, the maximum possible angle with the four cells configuration. The outer radius $r_o$ is also set to it's maximum allowable length, $r_{max}$ of 75 mm, limited by the geometric constraints of the robot. 

While the previous parameters are optimized without any trade-off, the inner radius $r_i$ is a more complex parameter to set. Indeed, the optimal inner radius \textcolor{modif}{shows a trade-off} between sensitivity, robustness to parasite motion, linearity and allowable stretch of the polymer film. On one hand, increasing the inner radius, increases the upper base $b_u$, improving sensitivity to in-plane translations (see eq. \eqref{eq:dclin}). But on the other hand, this decreases $h_0$, which will increase the sensor's non-linear behavior (see eq. \eqref{eq:sens2}), its sensitivity to parasite motion (see eq. \eqref{eq:tz} and eq. \eqref{eq:rxy}) and film stretch.

The allowable stretch $\lambda$ of the polymer film is limited by buckling (occuring when $\lambda < 1$) and a maximum stretch $\lambda_{max}$ at film failure, which is material dependant. It is thus necessary to satisfy:
\begin{equation}
1 \leq \frac{h_0 \pm y}{l_0} \leq \lambda_{max}  \quad \forall \; y
\label{eq:strain}
\end{equation}
where $l_0$ is the initial length of the film before pre-strech. This shows that the choice of $h_0$ is coupled to the choice of the initial pre-strech ($\lambda_0 = h_0/l_0$) and must consider the range of displacement of the inner frame.


For the initial prototype, a conservative approach is taken, with a 3x3 pre-stretched film made of two layers of 0.5 mm thick acrylic tape (3M VHB), for robustness and ease of manufacturing, and a inner radius $r_i$ set to the minimum value of 20 mm. Final geometrical parameters are shown in TABLE \ref{tab:geo}. Assuming a constant dielectric constant $\epsilon_r$ of 3.2 for the acrylic film, the predicted base capacitance of one cells $C_0$ is 0.9 nF and the predicted differential measurement sensitivity is 66 $pF/mm$ on both DOF using a cell pair to independently sense each DOF.
\begin{table}[h!]%
\centering
\begin{tabular}{c c c c}
\hline \hline
 $b_l$  &  $b_u$  &  $h_0$ & $d_0$ \\ \hline
 30 mm  &  100 mm   &  55 mm  & 0.11 mm \\ \hline \hline
\end{tabular}
\caption{Prototype Geometrical Parameters}
\label{tab:geo}
\end{table}

%% file: sections/experiments.tex
\section{Experiments}
\label{sec:exp}

\subsection{Setup}

\begin{figure}[htbp]
	\centering
		\includegraphics[width=0.48\textwidth]{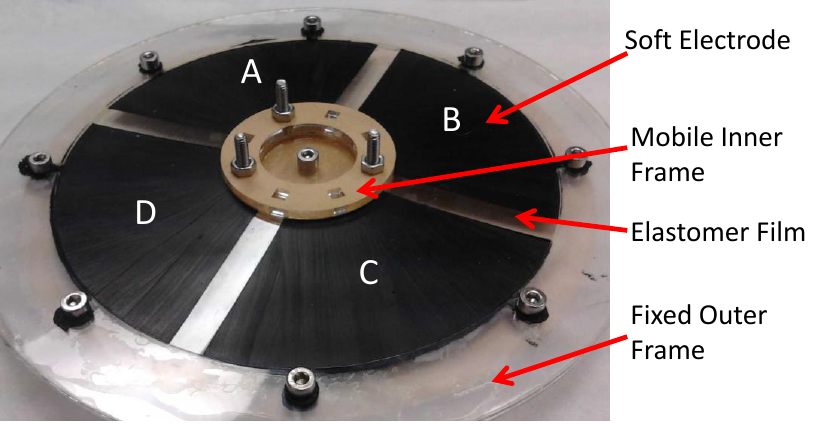}
	\caption{\textcolor{orange}{Prototype 2 DOF sensor with 4 electrode cells (A, B, C and D)}}
	\label{fig:proto}
\end{figure}

A full scale prototype of the 2DOF sensor, seen in Fig. \ref{fig:proto}, was manufactured. The \textcolor{orange}{pre-stret\textcolor{modif}{ch}ed dielectric} film was fixed to a pair of exterior and interior solid acrylic frames. The compliant electrodes, made \textcolor{orange}{from a custom mix of carbon powder in a silicone matrix (1g of carbon, 4g of silicone and 16g of kerosene)}, were hand painted \textcolor{orange}{on either side of the film} using laser-cut masks. 

\subsection{Protocol}

\begin{figure}[ht]
	\centering
		\subfloat[Static tests]{
  	\label{fig:jig_b}
  	\includegraphics[width=0.22\textwidth]{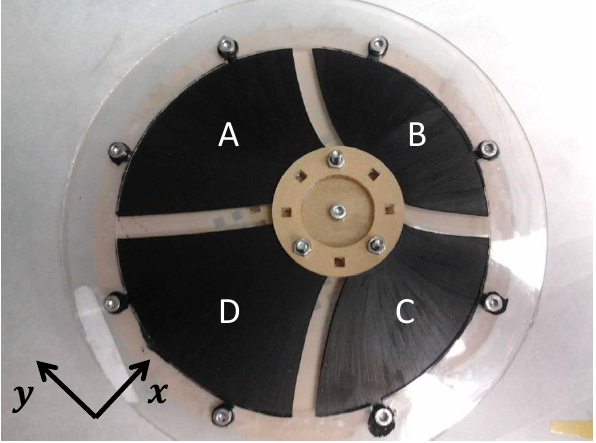}}
  	~
  	\subfloat[Jig for static tests]{
		\label{fig:jig_a}
		\includegraphics[width=0.22\textwidth]{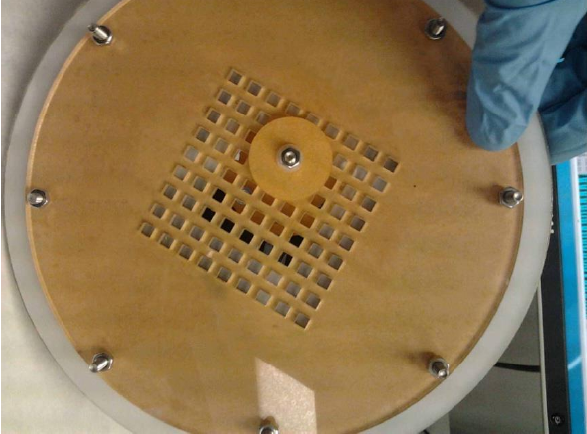}}
  	\caption{Static measurements setup \textcolor{orange}{using a 2D grid as a positioning jig}}
	\label{fig:jig}
	\vspace{-5pt}
\end{figure}

\textcolor{orange}{For the static tests, a jig was used to simulate different sensor positions, by manually locking the inner frame in discrete in-plane locations, as shown in Fig. \ref{fig:jig}.} All four sensing cells were sequentially measured using a custom capacitance reading circuit (detailed in sec. \ref{sec:capident}). Measurements were taken about 20 sec after positioning of the inner-frame to ensure complete relaxation of the \textcolor{orange}{viscoelastic} film. Three sets of positions were measured. The first two sets are meant to represent normal in-plane operation (the second set of measurements was performed with the positioning grid tilted \textcolor{modif}{at} a 22.5$^{\circ}$ angle with respect to the first set to reach new discrete positions). The third set was performed \textcolor{orange}{with a 3 mm} shim to create an out-of-plane translation along $z$ axis. Note that the setup did not allow rotation motion to be tested.

\subsection{Capacitance Identification}
\label{sec:capident}

The capacitance of the cells were estimated using a sensing algorithm developed by Gisby et al. \cite{gisby_dielectric_2012} that specifically compensates for the piezoresistivity of the electrodes as the cell deforms and the current induced due to the cell's rate of deformation \cite{gisby_smart_2011}. An oscillatory excitation signal was applied while simultaneously measuring the voltage across and current through the cell over a fixed period of time ($\sim$7ms), which was subsequently transformed according to the algorithm to derive capacitance using a least squares hyperplane fitting process. The measurement accuracy of the prototype system was limited by a noise level of $\sim$10pF. 

\subsection{Static Results}

\begin{figure}[htbp]
	\centering
		\includegraphics[width=0.48\textwidth]{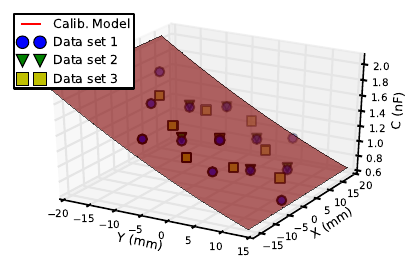}
	\caption{Capacitance of cell A \textcolor{orange}{in function of the inner frame displacement.}}
	\label{fig:C_3D}
\end{figure}

\begin{figure}[htbp]
	\centering
		\includegraphics[width=0.48\textwidth]{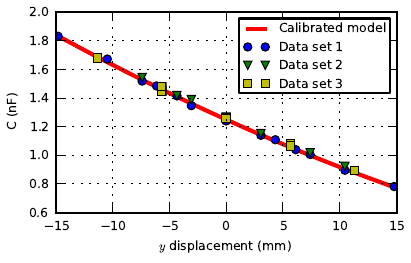}
	\caption{Capacitance of cell A \textcolor{orange}{(2D projection of Fig. \ref{fig:C_3D}).}}
	\label{fig:C_2D}
\end{figure}

Fig. \ref{fig:C_3D} shows one\textcolor{modif}{-}cell capacitance measurements, as a function of the inner frame's $x$ and $y$ translations. All points roughly fit a surface which is parabolic in the $y$ direction and constant in the $x$ direction, as demonstrated by its 2D projection on Fig. \ref{fig:C_2D}.

The capacitance of the \textcolor{modif}{one\textcolor{modif}{-}cell method} thus follows the predicted pattern, schematized in Fig. \ref{fig:strain}. The analytic equation for the surface, from eq. \eqref{eq:ch2}, can be expressed as a fonction of the initial capacitance $C_0$ and the initial height $h_0$ :
\begin{equation}
	C(x,y) = \frac{C_0}{h_0^2}(h_0+y)^2
\label{eq:C0H0}
\end{equation}
The initial capacitance of the cell $C_0$ is measured to be 1.250 nF on the prototype. Along with this initial measurement, the model requires only one additional calibration parameter: the initial height $h_0$. This is performed by minimizing the error between experimental data and model, yielding $h_0$ equal to 70 mm. The difference between this value and the true $h_0$ value of 55 mm, is explained by the rough trapezoid approximation of the cell shape.

\begin{figure}[htbp]
	\centering
		\includegraphics[width=0.48\textwidth]{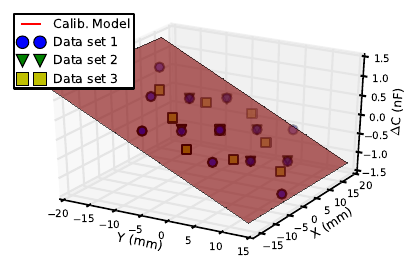}
	\caption{Delta capacitance of cell\textcolor{modif}{-}pair AC \textcolor{orange}{in function of the inner frame displacement.}}
	\label{fig:dC_3D}
\end{figure}

\begin{figure}[htbp]
	\centering
		\includegraphics[width=0.48\textwidth]{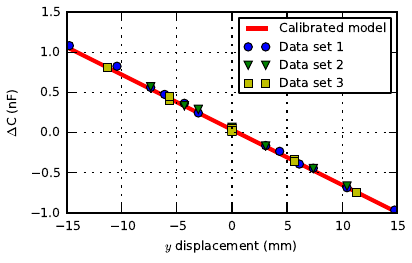}
	\caption{Delta capacitance of cell\textcolor{modif}{-}pair AC \textcolor{orange}{(2D projection of Fig. \ref{fig:dC_3D}).}}
	\label{fig:dC_2D}
\end{figure}

Fig. \ref{fig:dC_3D} shows the differential measurement results obtained between the pair of facing cells A and C. As predicted by eq. \eqref{eq:dclin}, the points fit on a plane; the capacitance difference is linear in the $y$ axis direction and constant in the $x$ axis direction, as illustrated by Fig. \ref{fig:dC_2D}, which is a projection of the Fig. \ref{fig:dC_3D}. \textcolor{orange}{Note that for both measurement approaches, the addition of an out-of-plane shim (data set 3), has a negligible effect; the data of set 3 fit the calibrated models as well as the data of set 1 and 2 (see Fig. \ref{fig:C_2D} and Fig. \ref{fig:dC_2D}).}

For each cell\textcolor{modif}{-}pair, the differential sensor calibration is done by fitting a plane on a data set using a least square criterion :
\begin{equation}
	\Delta C(x,y) = \alpha x + \beta y + \gamma
\end{equation}
TABLE \ref{tab:cal} shows the calibration results using all points.
\begin{table}[h!]%
\centering
\begin{tabular}{c c c c}
\hline \hline
 Cell\textcolor{modif}{-}Pair  &  $\alpha$                 &  $\beta$                 & $\gamma$           \\ \hline
 AC         &  0.5   $pF/mm$    &  -69.4  $pF/mm$  &  31  $pF$     \\ \hline
 BD         &  -69.5  $pF/mm$   &  -0.8   $pF/mm$  &  47  $pF$     \\ \hline \hline
\end{tabular}
\caption{Calibrated Plane Results}
\label{tab:cal}
\vspace{-15pt}
\end{table}
For one sensor, the calibrated planes are found very similar in shape, but with a 90$^{\circ}$ angle around the $z$ axis. The $\beta$ value of \textcolor{modif}{cell-pair} AC and $\alpha$ value of \textcolor{modif}{cell-pair} BD are very close to the predicted sensitivity. The $\alpha$ value of \textcolor{modif}{cell-pair} AC and $\beta$ value of \textcolor{modif}{cell-pair} BD are almost equal to zero, the theoretical value.  The small non-zero values can be explained by a small angular alignment offset between the jig grid and the sensor cells. If the cell\textcolor{modif}{-}pairs are well aligned with their axis, the inner frame displacement can be computed independently for both axes, using \textcolor{modif}{cell-pair} AC to compute the $y$ axis value and \textcolor{modif}{cell-pair} BD to compute the $x$ axis value.

Sensor accuracy is evaluated by computing the deviation between a set of data and a calibrated plane. TABLE \ref{tab:AC} shows the deviation results of all possible \textcolor{modif}{cell-pairs}. Note that results in grey (diagonal values) correspond to the deviation between a set of data and a plane that is calibrated with this same set. 

\begin{table}[htbp]
	\centering
		\begin{tabular}{ c c c c c }
		\hline
			                          &        &  \multicolumn{3}{c}{Error on  }   \\ 
			 Plane calibrated with   & Deviation  & Set 1   & Set 2   & Set 3 \\ \hline \hline \\
			 \multicolumn{5}{c}{ \textcolor{modif}{cell-pair AC}                      } \\ \hline \hline
			 Set 1                  & max    & \textcolor{gray}{0.55 mm} & 0.24 mm                   & 0.70 mm \\ 
			                        & RMS    & \textcolor{gray}{0.21 mm} & 0.16 mm                   & 0.26 mm \\ \\
			 Set 2                  & max    & 0.62 mm                   & \textcolor{gray}{0.20 mm} & 0.92 mm \\ 
			                        & RMS    & 0.29 mm                   & \textcolor{gray}{0.09 mm} & 0.26 mm \\ \\
			 Set 3                  & max    & 1.02 mm                   & 0.55 mm                   & \textcolor{gray}{0.38 mm} \\
			                        & RMS    & 0.27 mm                   & 0.13 mm                   & \textcolor{gray}{0.23 mm} \\ \hline \hline \\
			 \multicolumn{5}{c}{ \textcolor{modif}{cell-pair BD}                      } \\ \hline \hline
			 Set 1                  & max    & \textcolor{gray}{0.25 mm} & 0.28 mm & 0.49 mm \\ 
			                        & RMS    & \textcolor{gray}{0.13 mm} & 0.18 mm & 0.24 mm \\ \\
			 Set 2                  & max    & 0.47 mm                   & \textcolor{gray}{0.21 mm} & 0.64 mm \\ 
			                        & RMS    & 0.22 mm                   & \textcolor{gray}{0.14 mm} & 0.25 mm \\ \\
			 Set 3                  & max    & 0.46 mm                   & 0.43 mm                   & \textcolor{gray}{0.33 mm} \\ 
			                        & RMS    & 0.17 mm                   & 0.16 mm                   & \textcolor{gray}{0.22 mm} \\ \hline \hline  
		\end{tabular}
	\caption{Deviation between experimental data and calibrated models}
	\label{tab:AC}
	\vspace{-15pt}
\end{table}

\textcolor{orange}{\subsection{Preliminary bandwidth tests}}
\label{sec:dynamictests}

\begin{figure}[htbp]
	\centering
		\includegraphics[width=0.45\textwidth]{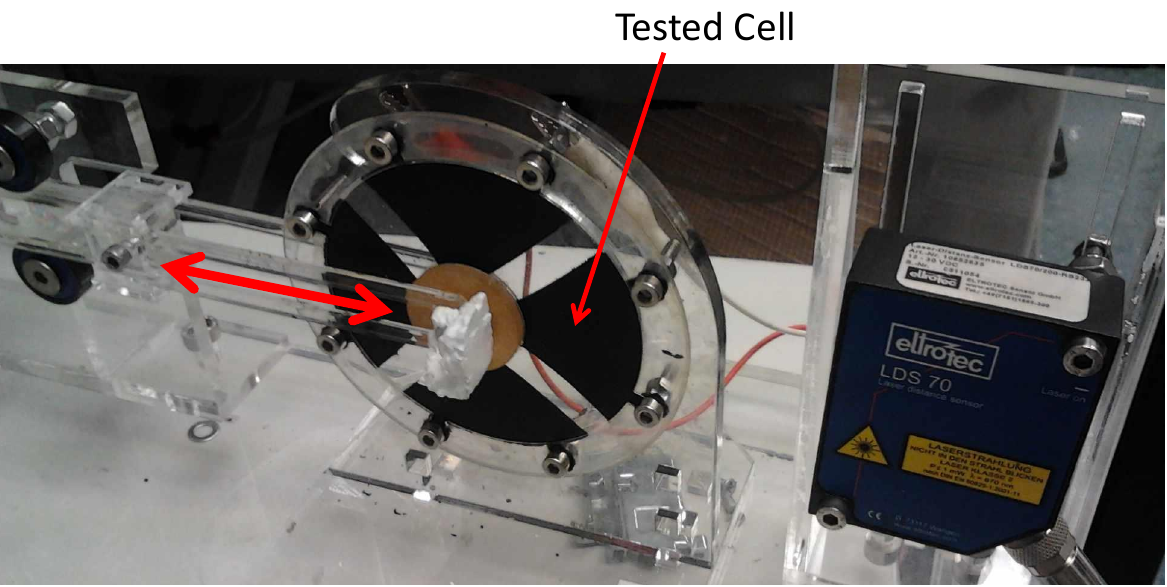}
	\caption{Dynamic measurements test bench}
	\label{fig:dyntestjig}
\end{figure}

\textcolor{orange}{Preliminary dynamic tests were also performed, with a 1:2 scale prototype (due to available equipment dimensions), in order to assess bandwidth limitations. The sensor was mounted on a test bench, see Fig. \ref{fig:dyntestjig}, with an actuator to impose 1D sinusoidal motion to the inner frame. A laser displacement sensor \textcolor{modif}{was used to measure} the inner frame displacement while the capacitance of \textcolor{modif}{the cell aligned with the imposed motion, i.e. having the highest sensitivity to displacements,} was simultaneously measured.}

\begin{figure}[htbp]
	\centering
		\includegraphics[width=0.48\textwidth]{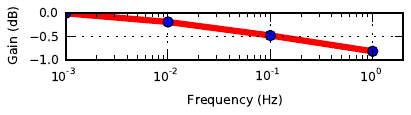}
	\caption{Experimental plot of capacitance over displacement gain.}
	\label{fig:bode}
\end{figure}

Four tests were conducted, with imposed sinusoidal motions at 0.001, 0.01, 0.1 and 1 Hz. The amplitudes of capacitance variation $A_C$ were compared to displacement amplitudes $A_y$ to compute the gain $G$ from displacement to capacitance:
\textcolor{modif}{
\begin{equation}
 G = \frac{ A_C}{A_y}
\end{equation}
}
\textcolor{modif}{Then, gains are expressed in decibel, taking the 0.001 Hz test as the 0 dB reference:}
\begin{equation}
 G_{db}=20\log_{10}\left(\textcolor{modif}{\frac{G}{G_{0.001 Hz}}}\right)
\end{equation}
Fig. \ref{fig:bode} shows the measured gain \textcolor{modif}{$G_{db}$} as a function of motion frequency.

\textcolor{orange}{The soft robotic system, for which the sensor is developed, is a point-to-point device and does not \textcolor{modif}{require} position feedback faster than one point each 5 sec. However, Fig. \ref{fig:bode} shows that even at such slow speeds the visco-elastic behavior of the acrylic film can produce errors in the order of a few percent. \textcolor{modif}{To overcome such limitations, silicone films could be used as their viscoelasticity is more than an order of magnitude less than that of acrylics} \cite{kornbluh_ultrahigh_2000}. The next generation of device will be made of silicone films and a more comprehensive dynamic study will be conducted to characterize their bandwidth.} 

%
%
%


%% file: sections/discussion.tex
\section{Discussion}
\label{sec:discussion}

Results show that the proposed linear model of the differential measurement approach captures the first order behavior of the sensor. However, parameter calibration is necessary to obtain good absolute precision since the dielectric constant $\epsilon_r$ of the film and the geometric values are hard to produce precisely. Fortunately, the proposed plane-like capacitance function allows a simple and convenient calibration process. This approach could be performed systematically after the manufacturing process, like it is done for the vast majority of commercially available sensors. TABLE \ref{tab:AC} shows that the calibrated model offers very good prediction capability. For the 12 cases that represent true deviations (not in grey), a mean RMS deviation of 0.2 mm is calculated, with a maximum deviation of 1 mm. The 0.2 mm RMS deviation represent only 0.7 \% of the sensor's full stroke, which is remarkable for a hand-manufactured sensor whose material cost would be well below 1 USD in mass production.

Using a custom sensing circuit, the noise level of the capacitance measurements was about $\pm$ 10 pF, which translates into $\pm$ 0.14 mm uncertainty when converted in displacement, by using the sensitivity of the prototype. This is very close to the RMS deviations of calibrated models, and does not even account for uncertainties caused by the manual positioning grid. We can thus conclude that the precision of the prototype was mainly limited by factors other than the sensor itself. 

\begin{figure}[htbp]
	\centering
		\includegraphics[width=0.40\textwidth]{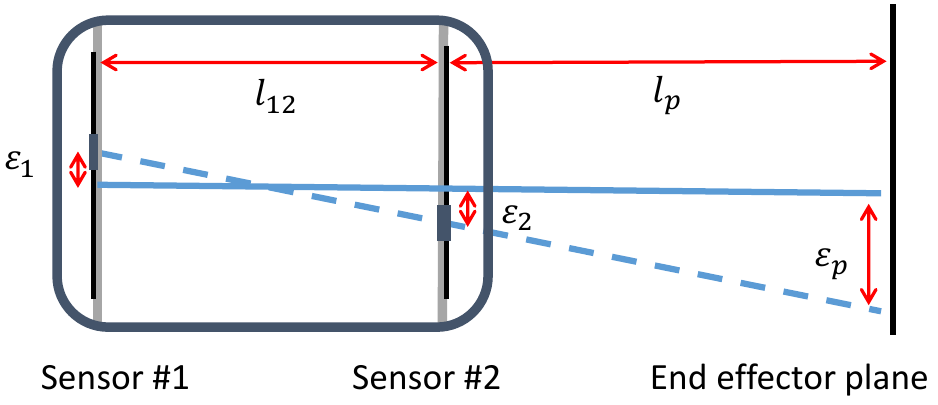}
	\caption{End-effector extrapolation \textcolor{orange}{of the error}}
	\label{fig:endeffector}
\end{figure}

Going back to the intended application, using the two sensor configuration, it is possible to extrapolate the final precision at the end-effector plane, see Fig. \ref{fig:endeffector}. 

The error propagation is given by :
\begin{equation}
	\frac{l_p}{l_{12}} \epsilon_1 + \frac{l_p+l_{12}}{l_{12}} \epsilon_{2} = \epsilon_p
\label{eq:error}
\end{equation}
Assuming normal independent distributions of errors and the same standard deviation on both sensors $\sigma_1 =\sigma_2 = \sigma$ , the standard deviation of the error on the end-effector plane is given by :
\begin{equation}
	\sigma_p = \sigma \sqrt{ \left( \frac{l_p}{l_{12}} \right)^2 + \left( \frac{l_p+l_{12}}{l_{12}} \right)^2 } 
\label{eq:errorrms}
\end{equation}
Using the geometrical values of the robot, $l_p$ equal to 200 mm and $l_{12}$ equal to 200 mm, and using the experimental RMS value as the standard deviation for both planes, a standard deviation of 0.45 mm at the end effector is found (on each translation DOF). This value meets the 1 mm requirement. Using a worst case scenario instead of the probabilistic assessment, i.e. using the maximal deviation of the worst case (model calibrated with the set 3 compare to the set 1) and assuming this unlikely event occur simultaneously on both sensors in opposed directions, then the error on the end-effector would reach 3 mm. 

\textcolor{modif}{Finally, the softness of the sensor was experimentally evaluated. A maximum  stiffness of 0.4 $N/mm$ was found in the radial direction. Hence, two planar sensors would not significantly contribute to the global stiffness of the robot. The stiffness of each air muscle is about 6 $N/mm$, and the sensors would thus contribute to less than 5\% of the global stiffness of the full system, on all DOF.}

Overall, the obtained sensor's accuracy meets the targeted application requirement for point-to-point operation. However, for closed-loop motion control, it would be desirable to have a sensor accuracy that is an order of magnitude more accurate than the desired overall system accuracy. A less noisy capacitance measurements and better positioning jig during calibration would be the first steps to improve accuracy, but there \textcolor{modif}{are several} ways to improve on the mechanical behavior of the sensor itself. First, using a less viscous more elastic material, such as silicone, \textcolor{modif}{would enable a higher bandwidth, translating into better dynamic performance}. \textcolor{modif}{Also,} silicone films have higher dielectric constant and can be made very thin using spin coating process, thus enabling high sensitivity designs. However, a downside of silicone films is the\textcolor{modif}{ir} lower maximum stretch ($\sim$2x2). Second, using a single layer film instead of a double layer of acrylic film would double the device sensitivity. Third, increasing the pre-stretch ratio from 3x3 to 4x4 would also double the sensitivity, but bringing the maximum strain closer to the film limit (6x6) and decreasing the film life. \textcolor{modif}{Fourth, a device using multiple layer electrodes would have its base sensitivity multiplied by the number of layers.}

%% file: sections/conclusion.tex
\section{Conclusion}

In this paper a 2 DOF dielectric elastomer soft sensor is proposed as a position feedback device for an intra-MRI soft polymer robot \textcolor{modif}{for surgical interventions}. An analytical model is developed to analyse the sensor's behavior and the influence of design parameters. A differential measurement approach and a four\textcolor{modif}{-}cell planar sensor design are proposed by taking into account sensitivity, linearity and robustness towards parasitic motions. A sensor prototype with a conservative and simple design is build and tested. Results corroborate model predictions, such as the highly linear behavior \textcolor{modif}{over the sensor's whole} operating range. Using a simple calibration method, a 0.2 mm \textcolor{modif}{accuracy} is achieved on both in-plane DOF, over a 30 mm x 30 mm range, resulting in an overall \textcolor{modif}{accuracy} of 0.7 \% per DOF. Moreover, this \textcolor{modif}{accuracy} could be easily and greatly improved by tweaking the geometrical parameters, using multi-layer electrodes and improving the noise to signal ratio of the electronic circuit. This demonstration shows that DE sensors have a great potential for accurate multi-DOF sensing. DE sensors also have many major advantages over traditional sensing technologies: softness, very low cost, high strain capability, and embedded multi-DOF design. Soft robotics system could greatly benefit from such devices. 

%

%% file: sections/ending.tex
\section*{Acknowledgment}


The experimental work for this study was performed at the Biomimetics Lab of the Auckland Bioengineering Institute, University of Auckland. The authors would like to thank the other members of the Biomimetics team for their help, especially Mahdieh Nejati, Antoni Harbuz and Samuel Schlatter.